\documentclass[runningheads]{llncs}

\usepackage{amssymb}
\usepackage{eccv}

\usepackage{eccvabbrv}

\usepackage{graphicx}
\usepackage{booktabs}
\usepackage[accsupp]{axessibility}  
\usepackage{lmodern}
\usepackage{enumitem}
\usepackage[most]{tcolorbox}
\usepackage{fvextra} 
\tcbset{
  promptbox/.style={
    colback=gray!10,
    colframe=gray!50,
    boxrule=0.5pt,
    arc=3pt,
    left=4pt,
    right=4pt,
    top=4pt,
    bottom=4pt,
    fonttitle=\bfseries,
    enhanced,
    breakable,
  }
}

\usepackage{hyperref}

\usepackage{orcidlink}

\begin{document}

\title{Learning Consistent Temporal Grounding between Related Tasks in Sports Coaching} 

\titlerunning{Consistent Temporal Grounding for Sports Coaching}

\author{Arushi Rai \and
Adriana Kovashka
}
\authorrunning{A.~Rai and A.~Kovashka}

\institute{University of Pittsburgh, Pittsburgh PA 15260, USA \\
\email{arr159@pitt.edu, kovashka@cs.pitt.edu}}

\maketitle

\begin{abstract}
Video-LLMs often attend to irrelevant frames, which is especially detrimental for sports coaching tasks requiring precise temporal localization.
 Yet obtaining frame-level supervision is challenging: expensive to collect from humans and unreliable from other models. We improve temporal grounding without additional annotations by exploiting the observation that related tasks, such as generation and verification, must attend to the same frames. We enforce this via a self-consistency objective over select visual attention maps of tightly-related tasks. Using VidDiffBench, which provides ground-truth keyframe annotations, we first validate that attention misallocation is a meaningful bottleneck.  We then show that training with our objective yields gains of +3.0\%, +14.1\% accuracy and +0.9 BERTScore over supervised finetuning across three sports coaching tasks: Exact, FitnessQA, and ExpertAF, even surpassing closed-source models.
\end{abstract}

\section{Introduction}
\label{sec:intro}

\begin{figure}[t]
    \centering
    \includegraphics[width=0.75\linewidth]{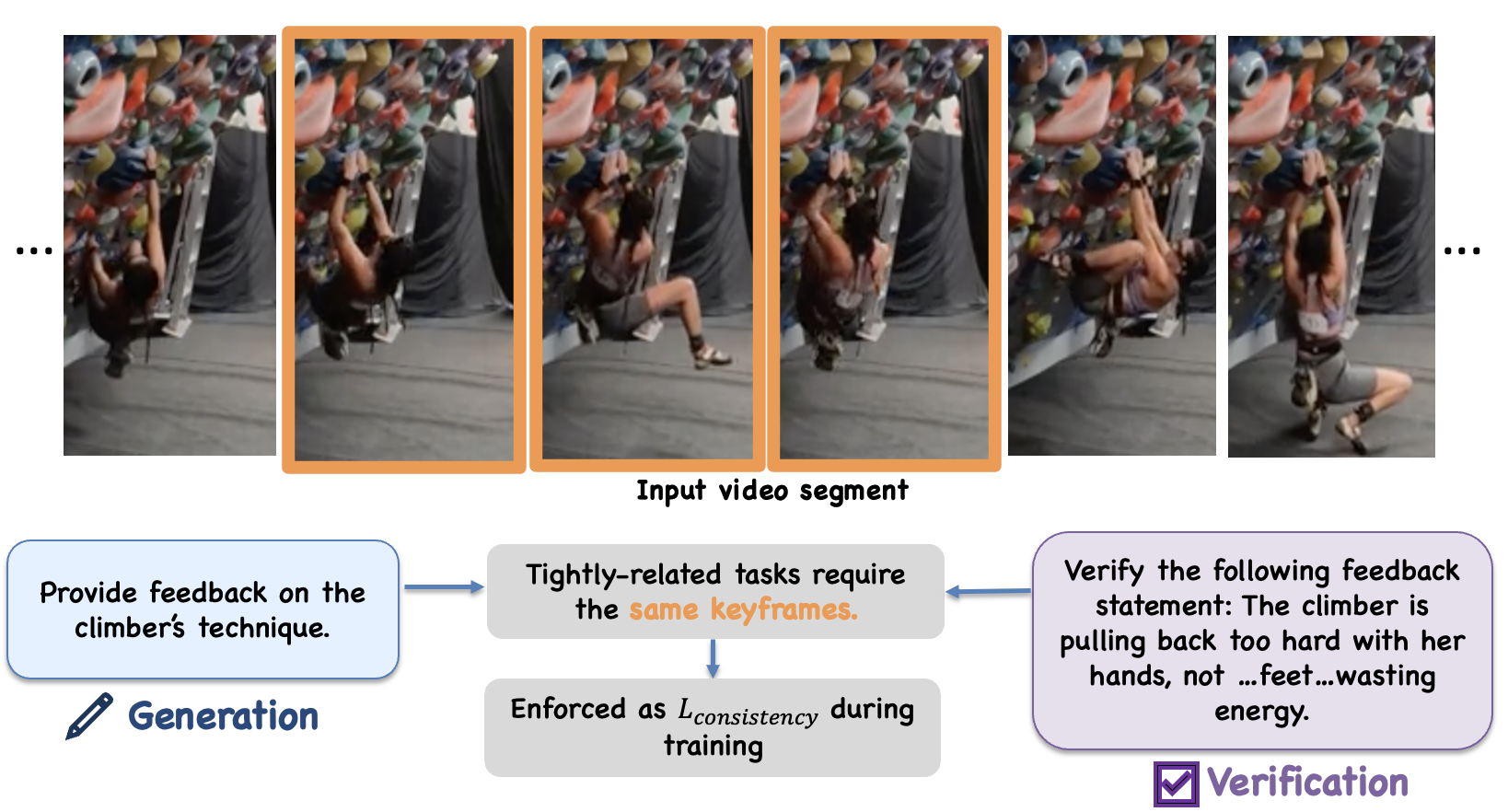}
\caption{Both the generation task (providing feedback on technique) and the verification task (confirming a given feedback statement) require the model to attend to the same keyframes (highlighted in orange) to produce a correct response. We exploit this as a self-supervision signal, enforcing attentional consistency between the two complementary task-views without requiring any frame-level annotations.}
    \label{fig:concept}
\end{figure}

Video question answering models often answer while attending to irrelevant frames, where correct answers may even rely on language priors or contextual cues rather than the direct visual evidence that actually supports the answer \cite{xiao2024can}. This failure mode is especially damaging for sports coaching tasks which require precise localization of frames to identify technique errors, assess form, or describe athletic actions, and where users must be able to trust that feedback is grounded in what actually happened. Consider the frames in Fig. \ref{fig:concept}; a poorly-grounded model asked to provide technique feedback on a climber mid-route would rely mostly on language priors and contextual cues, eliciting only the information that a person is climbing from the video. 
Such a model would produce common, plausible-sounding but often vague advice such as \textit{``Pull your hips closer to the wall''}, due to limited attention to the frames that actually reveal the problem. The true technique errors are visible only in the orange frames: the climber starts pulling up, without using proper foot positioning, and straining their arms under their body weight. A well-grounded model would attend to precisely these moments before generating feedback. 

Two approaches to tackle the challenge have been explored in the sports domain. The first collects real frame-level annotations \cite{burgess2025video}, but this is prohibitively expensive due to the large number of frames in a typical video, and is limited to a single benchmark, not a training dataset. The second, explored in concurrent work, uses pseudo-labels obtained from other models \cite{zou2025deepsport}, but risks propagating hallucinations about which frames actually matter. This motivates a self-supervised approach that encourages the model to attend to the right temporal evidence without requiring any additional frame-level annotations.

Self-supervised techniques in computer vision have successfully learned visual representations by exploiting invariances across different views of the same image, such as augmented crops, color jitter, or temporal frames, without requiring labels ~\cite{grill2020bootstrap, he2020momentum, chen2020simple}. 
Extending this idea to video-language models is non-trivial, as the input space is multimodal and the meaningful notion of a ``view'' is no longer obvious. 
One natural approach is to treat task prompt rephrasings as augmented views, encouraging the model to produce consistent representations across semantically equivalent inputs. However, rephrasings (e.g. ``What feedback would you give this athlete?'', ``Is this athlete's form correct? Yes or No'', or ``Summarize this video'') may fundamentally change the nature of the expected answer, and so enforcing hidden state or answer consistency is ill-defined. Additionally, even if answers or hidden states are consistent,  the model may attend to entirely different and potentially irrelevant frames in each rephrased task. Hence, output-level consistency does not guarantee temporal consistency.

We address the challenge of learning to attend to the right frames, by enforcing consistency 
at the level of visual attention maps, and using tightly-related tasks. 
We observe that tightly-related tasks, such as generating technique feedback and verifying whether a given feedback statement is correct, require the model to attend to the same keyframes regardless of differences in output form (shown in Fig.\ref{fig:concept}). 
While we do not know which frames these are, we use the observation about their \emph{sameness} as a self-supervised loss to guide the model's attention. 
We choose verification as guidance because it is more concrete (verify specific feedback claim) than the generation task where vague feedback and inaccurate attention are not penalized. 

Concretely, we route the same video through a shared model under two task prompts (i.e., a feedback generation pathway and a verification pathway). 
In select vision-centric layers and attention heads, we train the generation pathway to match the attention maps of the (simpler) verification pathway using a self-consistency loss.
This provides a direct, annotation-free training signal that encourages the model to localize the same temporal evidence regardless of the exact form of the task it is performing. 

To identify which layers and attention heads are most relevant for visual localization, we first conduct an analysis on VidDiffBench~\cite{burgess2025video} (a video difference benchmark with frame-level annotations) in Sec.~\ref{sec:analysis}. Using this dataset, we characterize visual attention quality across layers and heads, and validate that manipulating attention toward ground-truth keyframes (as proof-of-concept) improves performance. We then apply our self-consistency method, which \textbf{requires no frame-level annotations at training time}, and evaluate across a suite of sports coaching benchmarks: Exact~\cite{yi2025exact} (multiple-choice QA that evaluates expert-level understanding of physical skills), FitnessQA~\cite{panchal2024say} (fine-grained fitness coaching QA), and ExpertAF~\cite{Ashutosh2024ExpertAFEA} (expert-level sports feedback generation).

Our method consistently improves over the finetuned baseline across all benchmarks, and outperforms Gemini-3-Flash \cite{gemini3flash2026} on Exact and FitnessQA, which demonstrates that self-supervised temporal grounding can compensate for a substantial gap in model scale. 
To summarize, our contributions are:
\begin{itemize}[nolistsep,noitemsep]
    \item An analysis of visual attention quality across layers and heads in a video-language model, identifying attention misallocation as an independent bottleneck for sports understanding.
    \item A dual-pathway self-consistency training framework enforcing attentional agreement between complementary task views without frame annotations.
    \item Significant improvement (gains of $+3.0\%$ and $+14.1\%$ accuracy on Exact and FitnessQA, and $+0.9$ BERTScore on ExpertAF) over supervised finetuning in performance across multiple sports understanding benchmarks, even surpassing closed-source models.
\end{itemize}
\section{Related Works}
\textbf{Temporal grounding and visual localization in Vid-LLMs.}
A significant body of work improves temporal grounding in Vid-LLMs through architectural modifications \cite{rasekh2025enhancing, li2025improving}, chain-of-thought reasoning over frames \cite{zou2025deepsport, arnab2025temporal} or iteratively refining input \cite{wang2026timerefine, fan2025video, Hou2025}. While effective, these approaches require architectural changes, additional supervision, or increased inference cost. 

\textbf{Targeting attention distributions.}
A complementary direction targets the internal attention distributions of existing models directly, without modifying the architecture or requiring additional annotations. 
Attention exists as a mechanism for capturing relationships between tokens in transformer models \cite{vaswani2017attention}, making it a natural lens for explainability \cite{chefer2021generic} and, more recently, for diagnosing and correcting failures in multimodal LLMs. \cite{chen2025spatial} connects the persistent challenge of spatial reasoning in MLLMs to failures in visual attention, finding that uniformly-distributed attention is preferable to confidently concentrated but incorrect attention, and proposes a confidence-based intervention at inference time. 
\cite{kang2025see} similarly operates post-hoc, identifying image-centric attention heads and redistributing weight away from visual sink tokens that accumulate high attention without contributing meaningfully to the output. Post-hoc attention steering has also been explored via user-specified emphasis, both in language models \cite{zhang2023tell} and extended to the multimodal setting \cite{zhang2024prompt}. A related line of work uses attention to guide token pruning, retaining only the most relevant frames or tokens for downstream reasoning \cite{Hou2025}. Closer to our approach, \cite{kim2026compodistill} performs attention distillation from a larger teacher model to a smaller student, finding that attention misalignment is a significant bottleneck in transferring visual perception. Our method differs from all of the above in that it does not rely on post-hoc steering (although we test this in our analysis in Sec.~\ref{sec:analysis}), external supervision, or a separate additional teacher model. Instead, we enforce visual attention consistency between two task formulations of the same model, exploiting task invariance of visual attention as a self-supervised training signal.

\textbf{Sports understanding.} 
General-purpose MLLMs struggle on sports understanding tasks \cite{rao2025multi, xia2024sportu, gao2025fsbench, panchal2024say, burgess2025video, Ashutosh2024ExpertAFEA, he2025finebadminton}, in part due to the domain knowledge gap and spatio-temporal demands of the task. Several works have sought to close this gap by incorporating richer domain supervision or domain-knowledge graphs \cite{rao2025multi, chen2025finequest, yang2025soccermaster, gao2025fsbench, xia2026sportr}. \cite{Ashutosh2024ExpertAFEA} introduces the task of generating actionable expert feedback from video, jointly training on feedback generation, expert video retrieval, and 3D pose generation. \cite{ashutosh2025learning} proposes to jointly train feedback generation and classifying transferable skill-attributes such as balance and hand positioning that generalize across sports.  
More relevant to our work, recent \textbf{concurrent} efforts have explored post-training 
approaches to improve sports reasoning and temporal localization. \cite{bao2025tennistv} shows that 
reinforcement learning post-training alone does not necessarily improve performance on tasks requiring temporal grounding when sport-specific domain knowledge is not needed, suggesting that RL primarily enhances general content understanding rather than temporal grounding. This is further supported by DeepSport~\cite{zou2025deepsport}, a sports-specific reasoning model that uses chain-of-thought traces from 
open-source models~\cite{yang2025qwen3} for explicit temporal localization,
yet has limited improvement on assessment and coaching questions compared to recognition tasks (rule application, foul detection, action recognition) despite localization-specific RL post-training. DeepSport's error analysis of the trained model reveals frequent hallucination of temporal timestamps on fine-grained tasks, suggesting that post-training alone or distilled traces are insufficient to ground the model's attention in the correct frames. SportR~\cite{xia2026sportr} similarly finds through error analysis that visual perception and hallucination comprise the majority of errors from their sports reasoning model. Motivated by this, our method targets temporal grounding  directly through self-supervised attention consistency, without relying on  any chain-of-thought supervision or frame-level annotations, and shows consistent gains on assessment and coaching tasks~\cite{Ashutosh2024ExpertAFEA, yi2025exact, panchal2024say}. Our method could be considered complementary to CoT supervision and post-training.


    \section{Method}
    \label{sec:method}
    
    \begin{figure}[t]
    \centering
    \includegraphics[width=0.8\linewidth]{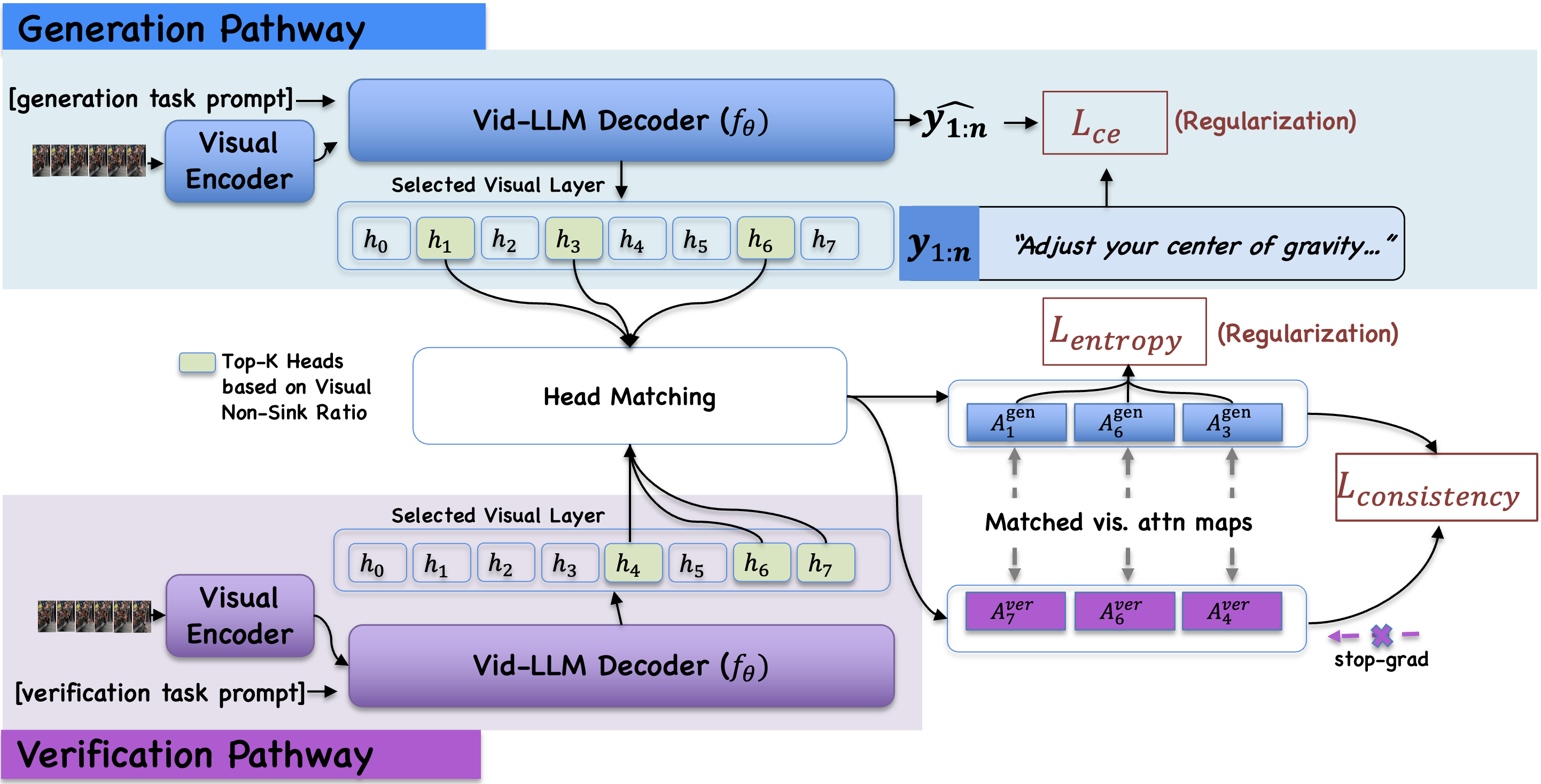}
    \caption{Overview of the proposed dual-pathway self-consistency framework.  The \textbf{Generation Pathway} and \textbf{Verification Pathway} process 
    the same video through a shared visual encoder and Vid-LLM Decoder ($f_\theta$) with different task prompts. From a selected visual layer (Sec.~\ref{sec:analysis}), 
    top-$K$ attention heads are identified via their Visual Non-Sink Ratio~\cite{kang2025see} and paired across pathways via \textbf{Head Bipartite Matching} (exact index 
    matching + Hungarian algorithm). The generation pathway's attention maps are regularized by $\mathcal{L}_{entropy}$ for focused attention over specific visual tokens (and frames), while $\mathcal{L}_{consistency}$ enforces cross-pathway attentional agreement between matched heads. $\mathcal{L}_{ce}$ serves as regularization on the 
    generation pathway output.}    
    \label{fig:method}
\end{figure}

We propose a dual-pathway self-consistency framework for  video-language model (Vid-LLM) training that improves internal visual localization \textbf{without requiring any external grounding supervision}. Self-attention in middle LLM decoder layers localizes task-relevant visual tokens~\cite{zhang2025cross}, making them key into how the model grounds its responses. These select attention modules should attend to consistent visual regions for two tightly related tasks. 
We exploit this by routing the same video through two task-conditioned pathways: a \textbf{generation pathway} (harder task) and a \textbf{verification pathway} (easier task) through a shared model, and enforcing attentional consistency between their internal attention maps. 
Figure~\ref{fig:method} provides an overview. 
If the model is genuinely grounded, both tasks should localize the same visual regions, and encouraging this agreement provides a useful inductive bias that discourages spurious or task-inconsistent visual attention. 
    
    \subsection{Preliminaries}
    \label{sec:prelim}
    
    Let $x^{vis} = \{x^{vis}_1, \ldots, v^{vis}_T\}$ denote a sequence of $T$ sampled video frames. A Vid-LLM processes $\mathcal{V}$ via a visual encoder $f_{\phi}$ to produce frame-level visual tokens, which are concatenated with a text prompt $x^{text}$ and passed to an autoregressive decoder $f_{\theta}$ to generate a response $\hat{y}_{1:n}$. Formally:
    \begin{equation}
        \hat{y}_{1:n} = f_\theta\bigl(f_{\phi}(x^{vis}),\, x^{text}\bigr).
    \end{equation}

    \subsection{Dual-Pathway Forward Pass}
    
    Given a video $x^{vis}$ and a paired generation-verification prompt pair $(p^{\text{gen}}, p^{\text{ver}})$, we perform two forward passes through the shared frozen encoder $f_{\phi}$ and trainable decoder $\theta$.
    
    \noindent \textbf{Generation Pathway.}
    The generation pathway takes the video $\mathcal{V}$ and a generation task prompt $p^{\text{gen}}$ (e.g., a sports feedback \cite{Ashutosh2024ExpertAFEA} or sports understanding QA instruction \cite{panchal2024say, yi2025exact}) and produces predicted output tokens $\hat{y}_{1:n}$, supervised by a standard cross-entropy loss:
    \begin{equation}
        \mathcal{L}_{ce} = -\sum_{t=1}^{n} \log \theta\bigl(y_t \mid y_{<t},\, f_{\phi}(x^{vis}),\, p^{\text{gen}}\bigr).
    \end{equation}
    Within our self-consistency training framework, this loss serves as a regularization term that preserves the model's generation capability.
    
    \noindent \textbf{Verification Pathway.}
    The verification pathway processes the same video $\mathcal{V}$ with a verification task prompt $p^{\text{ver}}$ through the same shared decoder $\theta$. 
    The verification task prompt is constructed by combining the ground-truth answer $y_{1:n}$ with an automatically rephrased version of the generation prompt $p^{\text{gen}}$, forming a yes/no question that asks whether $y_{1:n}$ is a valid response to the video given the rephrased prompt (e.g. ``Please verify if the following feedback statement is correct for the given video: {answer}''). \textbf{No gradients flow through the verification pathway}, to prevent training collapse \cite{caron2021emerging, grill2020bootstrap, he2020momentum}. 
    The verification pathway contributes to training exclusively through the self-consistency loss $\mathcal{L}_{consistency}$ applied to its attention maps, with the stop-gradient operator (Sec.~\ref{sec:losses}) ensuring that the verification pathway acts purely as a target signal.

    \subsection{Visual Layer and Head Selection}
    
     The choice of both heads and layer is critical, as not all heads and layers encode meaningful temporal information equally. 
     We adopt different selection strategies for each. For the \emph{layer}, we are motivated by prior findings in \cite{zhang2025cross}, which suggests that the layers most responsible for encoding task-specific visual information remain \textbf{consistent} across similar tasks. We choose $\ell^*$ using the visual attention quality score (Eq.\ref{eq:visionquality}) which captures criteria of sharpness, vision-centricity, and keyframe overlap that intuitively align with temporal grounding potential 
     and keep this fixed (manually selected) for the model. 
     We discuss this in more detail and provide empirical evidence in Sec.~\ref{sec:analysis}.
     We found that manual (i.e. selecting based on the visual attention quality score) \emph{head} selection, once the layer is selected manually, is suboptimal empirically. 
     We therefore select the top-$K$ heads dynamically per forward pass using the Visual Non-Sink Ratio\cite{kang2025see} ($\mathrm{VNSR}$),
     which measures how much visual attention falls on non-sink versus sink tokens (see supp for full definition). Since visual sink tokens attract disproportionately high attention due to large feature-dimension activations despite carrying no task-relevant content, $\mathrm{VNSR}$ favors heads whose attention is concentrated on meaningful visual content. 
    We select the top-$K$ heads from each pathway ranked by $\mathrm{VNSR}$, yielding head sets $\mathcal{H}^{\text{gen}} = \{h^{\text{gen}}_1, \ldots, h^{\text{gen}}_K\}$ and $\mathcal{H}^{\text{ver}} = \{h^{\text{ver}}_1, \ldots, h^{\text{ver}}_K\}$. We use $K=5$, and ablate in exp (Sec.\ref{sec:ablations}).
    
    \subsection{Head Bipartite Matching}
    
    To enforce self-consistency, we first must establish a correspondence between heads in $\mathcal{H}^{\text{gen}}$ and $\mathcal{H}^{\text{ver}}$. We propose \textbf{Head Bipartite Matching}, a two-stage procedure that combines exact index matching with the Hungarian algorithm \cite{hungarian} for assignment.
    
    \noindent \textbf{Stage 1: Exact Index Matching.}
    We first match heads that share the same index across both pathways, i.e., $h^{\text{gen}}_i \leftrightarrow h^{\text{ver}}_i$ if $i \in \mathcal{H}^{\text{gen}} \cap \mathcal{H}^{\text{ver}}$ within the selected visual layer.
    
    \noindent \textbf{Stage 2: Hungarian Matching.}
    For remaining unmatched heads, we compute a pairwise cost matrix $\mathbf{C} \in \mathbb{R}^{K' \times K'}$ between the unmatched heads of each pathway, where $K'$ is the number of remaining heads. 
    The cost between head $i$ from the generation pathway and head $j$ from the verification pathway is the asymmetric KL divergence $D_{\text{KL}}( \mathbf{A}^{\text{ver}}_j \| \mathbf{A}^{\text{gen}}_i)$, computed over visual token indices $\mathcal{I}_{\text{vis}}$, consistent with the directional formulation of $\mathcal{L}_{\text{consistency}}$ (Eq.~\ref{eq:kl}). The Hungarian algorithm then finds the optimal 1:1 assignment permutation $\sigma^*: \{1,\ldots,K'\} \to \{1,\ldots,K'\}$, and the combined matching from both stages yields a full set of $K$ matched head pairs $\{(h^{\text{gen}}_i, h^{\text{ver}}_{\sigma^*(i)})\}_{i=1}^{K}$.
    
    \subsection{Losses}
    \label{sec:losses}
    Let $\mathbf{A}^{\text{gen}}_i$ and $\mathbf{A}^{\text{ver}}_i$ denote the visual attention maps produced by the $i$-th matched head pair from the generation and verification pathways, respectively, where each map $\mathbf{A_i} \in \mathbb{R}^{1 \times N_{\text{vis}}}$ captures the attention from the last prompt token to all visual tokens, $N_{vis}=|\mathcal{I}_{\text{vis}}|$.
    
    \noindent \textbf{Entropy Regularization.}
    To encourage the generation pathway's attention heads to produce focused, peaky attention distributions over visual tokens rather than uniform or sink token-dominated maps, we apply an entropy regularization loss over the selected vision-centric heads:
    \begin{equation}
        \mathcal{L}_{entropy} = \frac{1}{K} \sum_{i=1}^{K} \mathcal{H}\bigl(\frac{\mathbf{A}^{\text{gen}}_i}{\tau_{entropy}}\bigr),
    \end{equation}
    where $\mathcal{H}(\cdot)$ denotes the Shannon entropy computed over the visual token dimension and $\tau_{entropy}$ scales the distribution to be sharper.
    
    \noindent \textbf{Self-Consistency Loss.}
    
    \begin{equation}
        \mathcal{L}_{consistency} = \frac{1}{K} \sum_{i=1}^{K} D_{\text{KL}}\Bigl(\text{sg}\bigl(\mathbf{A}^{\text{ver}}_i \,\Big\|\, \mathbf{A}^{\text{gen}}_i \bigr)\Bigr),
        \label{eq:kl}
    \end{equation}
    where $\text{sg}(\cdot)$ denotes the stop-gradient operator. We use KL divergence as it is a natural measure of distributional discrepancy between attention distributions. The asymmetric direction treats the verification pathway as a fixed reference toward which the generation pathway is trained. The stop-gradient prevents training collapse, analogous to the fixed target network in BYOL~\cite{grill2020bootstrap} and MoCo~\cite{he2020momentum} in self-supervised visual representation learning methods. \textbf{Additionally, since verification is an easier task than generation, its attention maps provide a more reliable target signal.}
    
    \noindent \textbf{Total Training Objective.}
    The final training objective combines all three losses with equal weighting:
    
    \begin{equation}
        \mathcal{L} = \mathcal{L}_{ce} + \mathcal{L}_{entropy} + \mathcal{L}_{consistency}
    \end{equation}
    
    We find this equal weighting to be robust across our experimental settings, but future work could add additional hyperparameters.
    
\section{Experiments}




To precisely apply our method to modules related to temporal grounding, we define heuristics and identify suitable candidate layers and heads in Sec.~\ref{sec:analysis}. We then verify these modules' sensitivity to temporal grounding quality by redistributing visual attention toward ground-truth keyframes and away from irrelevant frames in Sec.~\ref{sec:manipulating_attn_distribution}.  We additionally test strategies to adaptively select heads for redistribution, which we use in our method. Results from improving based on the layer selection and dynamic head selection strategy inform which modules to apply our self-consistency and entropy regularization loss. We show gains on all three datasets in Sec.~\ref{sec:main_results} and ablate each component in Sec.~\ref{sec:ablations}. 

\textbf{Experiment Details.} All experiments use 1 GPU with an aggregated batch size of 4, implemented via gradient accumulation over 4 steps with a per-step batch size of 1. We use a learning rate of $2\times10^{-4}$ with LoRA (rank 16, alpha 32, dropout 0.05) applied solely to the language model. We set $\tau_{entropy}$ to 0.03, to significantly sharpen the internal visual attention due to very high initial entropy. We use PyTorch with PerceptionLM \cite{cho2025PerceptionLM} as our base model, with the visual pooling rate (controls \# of tokens per frame in PerceptionLM \cite{cho2025PerceptionLM}) as 5 and uniformly sample 16 frames during training due to GPU memory constraints. During inference, visual pooling rate is set to 4 for a richer visual representation, and the sampling temperature is set to 1.

\textbf{Datasets.} We select three sports coaching datasets that test fine-grained understanding and feedback generation. Exact \cite{yi2025exact} and FitnessQA \cite{panchal2024say} both evaluate fine-grained sports coaching understanding, with questions targeting technique quality, cause-effect of motions, and similar aspects. Exact is drawn from EgoExo4D \cite{Grauman2023EgoExo4DUS}, covering physical skilled demonstrations across sports ranging from soccer to dance, and provides curated multiple-choice questions for fine-grained understanding. FitnessQA covers exercise demonstrations (squats, push-ups, etc.) with fine-grained questions about movement quality, but uses open-ended answers. Since open-ended answers require either an LLM-as-a-judge or imprecise semantic similarity metrics to evaluate, we use an LLM to convert the original annotations to multiple-choice questions (prompt and examples provided in supp), enabling straightforward accuracy-based evaluation. ExpertAF \cite{Ashutosh2024ExpertAFEA} is a feedback generation task, where the model must produce open-ended coaching feedback rather than answer specific questions about technique, making it unsuitable for multiple-choice conversion. We therefore evaluate predicted feedback against reference text using BERTScore \cite{zhang2019bertscore}. 
We train on 800 samples each from Exact \cite{yi2025exact}, FitnessQA \cite{panchal2024say}, and ExpertAF \cite{Ashutosh2024ExpertAFEA}. We evaluate on the 200-sample test split of Exact \cite{yi2025exact}, 659-sample test split of FitnessQA \cite{panchal2024say}, and the entire test-split of ExpertAF (7K samples).

\begin{figure}[t]
    \centering
    \begin{subfigure}[t]{0.4\linewidth}
        \centering
        \includegraphics[width=\linewidth]{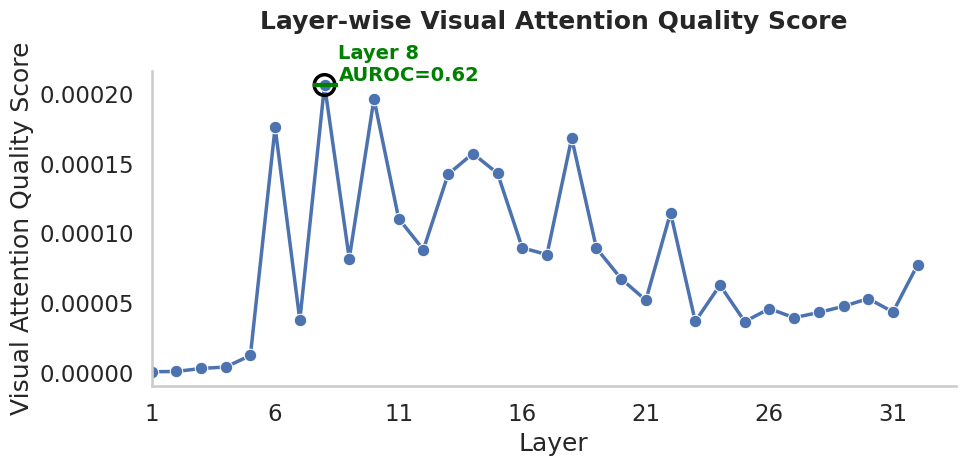}
        \caption{Layer-wise visual attention quality score.}
    \end{subfigure}
    \hfill
    \begin{subfigure}[t]{0.40\linewidth}
        \centering
        \includegraphics[width=\linewidth]{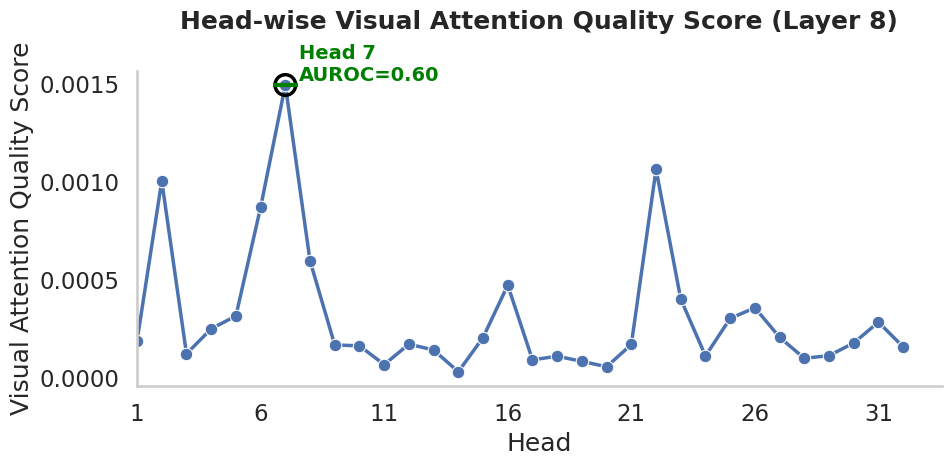}
        \caption{Head-wise visual attention quality score.}
    \end{subfigure}
            \caption{Visual attention quality score visualization: product of attention sharpness ($1-\text{entropy}$), vision centricity (visual attention sum), and keyframe overlap (AUROC). Both graphs show that layer 8 and head 7 are relatively better at temporal grounding.}
    \label{fig:auroc-viddiff-bench}
\end{figure}

\subsection{Which attention modules are sensitive to temporal grounding?}
\label{sec:analysis}
Before validating our method, we first seek to understand which attention modules selectively attend to tokens from relevant frames, and then later in \cref{sec:manipulating_attn_distribution} test if they can be perturbed to improve sports coaching performance.
Prior work has explored the characteristics of these attention distributions within the image context of MLLMs \cite{zhang2025cross, zhang2025mllms, chen2025spatial} finding that certain layers and heads are more vision-centric than others, and that similar mid-level layers tend to specialize in visual grounding consistently across similar tasks. However, since this work only examines spatial grounding in images, it is unclear whether the same patterns hold for temporal grounding in video. To find which layer is a good candidate for \textbf{improving} temporal grounding in the video domain, 
we use the EgoExo4D \cite{Grauman2023EgoExo4DUS} split of VidDiffBench \cite{burgess2025video}, which provides ground-truth keyframe annotations per question; these are all known to be relevant. 

\textbf{Defining temporal grounding sensitivity.} A layer may be sensitive to temporal grounding yet still attend to the wrong frames, meaning it has the capacity to ground but is currently misaligned. To identify such layers, we propose heuristics that characterize attention quality along two axes, in addition to current temporal grounding ability: (1) \textit{selectiveness}, measured by attention sharpness since select frames should contribute to the answer in this task; (2) \textit{vision-centricity}, which measures how much attention mass is allocated to visual tokens, compared to text or system tokens; and (3) \textit{keyframe overlap}, measured by the AUROC between the attention distribution over visual tokens and the ground truth keyframe indicator; a similar characterization for objects within a single image was proposed in \cite{chen2025spatial}. We combine these three axes into a \textbf{single visual attention quality score}:

\begin{align}
    S(\bar{A^\ell}) = \underbrace{\left(1 - \frac{H(\bar{A^\ell})}{\log N_{\text{visual}}}\right)}_{\text{selectiveness}} + \underbrace{\sum_{i \in \mathcal{V}} \bar{A_i^\ell}}_{\text{vision-centricity}} + \underbrace{\text{AUROC}(\bar{A^\ell}_{\mathcal{V}},\, \mathbf{k})}_{\text{keyframe overlap}}
    \label{eq:visionquality}
\end{align}
where $\bar{A^\ell}\in \mathbb{R}^{N}$ is the mean attention vector at layer $\ell$, 
averaged over all heads and samples, $\bar{A^\ell}_{\mathcal{V}} \in \mathbb{R}^{N_{\text{visual}}}$ 
is its restriction to visual token indices $\mathcal{V}$, $N_{\text{visual}}=|\mathcal{V}|$, and $\mathbf{k}  \in \{0,1\}^{N_{\text{visual}}}$ is the ground truth keyframe indicator where 
\begin{align}
\mathbf{k}_i = \begin{cases} 1 & \text{if token } i \text{ belongs to a keyframe} \\ 0 & \text{otherwise} \end{cases}
\end{align}
The entropy (for computing sharpness) is calculated over the visual attention distribution and captures its peakiness: 
\begin{align}
    H(\bar{A^\ell}) = -\sum_{i\in\mathcal{V}} p_i \log p_i, \quad p_i = \frac{\bar{A^\ell}_i}{\sum_{j\in\mathcal{V}} \bar{A^\ell}_j}.
\end{align}


\textbf{Layer and head manual selection analysis.} As shown in Fig.\ref{fig:auroc-viddiff-bench}(a), layer 8 achieves the highest visual attention quality score. However, the keyframe overlap (AUROC) is 0.62, where 0.5 means no better than random guessing, which reveals that the model's internal attention does not reliably localize ground truth keyframes, indicating that the bottleneck could lie in visual representation or attention allocation. When inspecting individual heads within layer 8 in Fig.\ref{fig:auroc-viddiff-bench}(b), head 7 yields the highest quality score, yet we find that its average attention is heavily concentrated on the first few visual tokens regardless of where the keyframes appear. We attribute this to the visual sink token phenomenon \cite{kang2025see}, where large feature-dimension activations disproportionately attract attention mass. Since prior work finds that vision-centric layers are consistent across similar tasks \cite{zhang2025cross}, we fix the layer to layer 8 and use adaptive head selection (described next) to favor heads not dominated by visual sink tokens for our method.

\begin{table}[t]
\centering
\caption{Proof-of-concept analysis: Effect of manipulating attention distributions using ground-truth \emph{keyframes} (relevant frames) on the EgoExo4D split of VidDiffBench~\cite{burgess2025video}. Biasing attention toward annotated keyframes improves accuracy over uniform or no redistribution, demonstrating the benefit of temporal (frame) grounding. Regarding head selection, automatic selection via visual non-sink ratio \cite{kang2025see} performs best.}
\begin{tabular}{l|l|c}
\toprule
\textbf{Head Selection} & \textbf{Redistribution} & \textbf{Accuracy} \\
\midrule
\multicolumn{3}{c}{\textit{Manual Head Selection (layer=8, head=7)}} \\ \midrule
\quad None          & None                          & 65.8 \\
\quad None          & Uniform over all frames                & 68.4 \\
\quad None          & Proportional over keyframes   & \textbf{71.8} \\
\addlinespace
\midrule
\multicolumn{3}{c}{\textit{Adaptive Head Selection (layer=8)}} \\\midrule
\quad Top-$K$ vis. attention sum   & Proportional over keyframes   & 69.2 \\
\quad Top-$K$ vis. non-sink ratio  & Proportional over keyframes   & \textbf{73.4} \\
\bottomrule
\end{tabular}
\label{tab:selection_results}
\end{table}
\subsection{Can temporal grounding improve performance independently of visual representation?}
\label{sec:manipulating_attn_distribution}
Our method targets attention misallocation as a meaningful bottleneck. In this section, we use ground-truth keyframe labels as a proof-of-concept that correcting attention misallocation improves performance, and to inform two design choices: fixing the layer and using adaptive head selection.

\textbf{Attention redistribution strategies.} To test if reallocating the visual attention based on keyframes (relevant frames) can improve performance, we compare attention reallocation against two baselines: no manipulation, and uniform redistribution over all visual tokens (without considering keyframes). Since modern MLLMs \cite{cho2025PerceptionLM, yang2025qwen3, wang2025internvideo} represent each frame with multiple tokens, we 
redistribute the visual attention \textit{proportionally over keyframes} which preserves the relative within-keyframe attention distribution (see figure in supp). As shown in top half of Table~\ref{tab:selection_results}, the proportional redistribution performs the best (71.8)  compared to uniform redistribution and no manipulation. This shows potential impact for our method: if the model learns to attend better in these temporal grounding sensitive modules using our method, performance will improve.

\textbf{Adaptive head selection.} We explore adaptive identification of the top-K vision-centric heads in layer 8, since the heads activated change per input (motivating the dynamic head selection in our method). We test using both the visual non-sink ratio \cite{kang2025see} or the visual attention sum to rank and select top-K heads ($K=5$). Selecting by visual non-sink ratio is more effective (73.4), as selecting by attention sum alone risks including sink-dominated heads that allocate high attention mass to tokens that might not be carrying task-relevant visual information. This informs our choice to select top-K heads using visual non-sink ratio in our method. Still, both of these head selection methods (and even static head selection) improve over no or uniform redistribution. This confirms that a large limitation is attention misallocation rather than solely model capacity or visual representation: correcting which frames the model attends to, even post-hoc, yields consistent gains, \textbf{motivating our method to do so without requiring ground-truth keyframe labels.}


\subsection{Does targeting temporal grounding sensitive modules with our method improve sports coaching performance?}
\label{sec:main_results}

\begin{table}[t]
\centering
\caption{Performance comparison across sports understanding benchmarks. Our method improves over the finetuned baseline on all evaluated datasets. * denotes discarding answer extraction failures}
\begin{tabular}{lccc}
\toprule
\textbf{Method} & \textbf{Exact} & \textbf{FitnessQA} & \textbf{ExpertAF} \\
\midrule
Gemini-3-Flash                          & 79.0 & 54.2   & -- \\
Qwen2.5-VL-7B-Instruct                          & --  & 33.3*   & 27.3 \\
Qwen3-VL-8B-Instruct                          & 40.2   & 55.4*  & 27.1 \\
\midrule
PerceptionLM-8B                 & 27.1 & 39.5 & 26.0 \\
PerceptionLM-8B (Finetuned)     & 84.5 & 74.1  & 37.5 \\
+Ours & \textbf{87.5} & \textbf{88.2} &  \textbf{38.4}\\
\bottomrule
\end{tabular}
\label{tab:main_results}
\end{table}
We evaluate our method on three sports coaching benchmarks and compare against several baselines. The most direct comparison is the finetuned PerceptionLM-8B baseline, 
and we additionally compare against the larger, closed-source Gemini-3-Flash \cite{gemini3flash2026} and two open-source Vid-LLMs, Qwen2.5-VL-7B-Instruct \cite{qwen25} and Qwen3-VL-8B-Instruct \cite{yang2025qwen3}.

As shown in Table~\ref{tab:main_results}, our method consistently improves over the fine-tuned PerceptionLM-8B baseline across all evaluated datasets. We observe gains of $+3.0$ on Exact, $+14.1$ on FitnessQA, and $+0.9$ BERTScore on ExpertAF, demonstrating that enforcing attentional self-consistency during training leads to meaningful improvements across tasks requiring precise localization under diverse output formats and evaluation criteria. Crucially, these gains are achieved without additional annotations, increased inference cost, or architecture modifications.

\textbf{Notably, an 8B open-source model can surpass closed-source systems on fine-grained sports understanding.} Our method achieves 87.5 on Exact and 88.2 on FitnessQA, outperforming Gemini-3-Flash (79.0/54.2) despite substantial differences in model scale and training data. While visual representation quality prior to the LLM decoder and overall model scale remain important factors, these results suggest that attention localization constitutes an independent and addressable bottleneck. With model capacity and visual representations held fixed, explicitly encouraging consistent temporal attention yields further performance gains.

\subsection{Ablation Studies}
\label{sec:ablations}

\begin{table}[t]
\centering
\caption{Ablation over head matching strategies and task for $\mathcal{L}_{\text{consistency}}$. 
``Fixed'' denotes that both pathways share the same selected layer $\ell^*$; 
``Flexible'' allows the verification pathway to select its own layer independently. ``Verification $\to$ Summarization'' replaces the verification pathway prompt with a summarization task.}
\setlength{\tabcolsep}{8pt}
\begin{tabular}{p{2.5cm} p{4.5cm} cc}
\toprule
\textbf{Verif.} \textbf{Layer} & \textbf{Head Matching Strategy} & \textbf{Exact} & \textbf{FitnessQA} \\
\midrule
Fixed ($\ell^* = 8$) & Ex. Match + Hung. (ours) & \textbf{87.5} & \textbf{88.2} \\
Fixed ($\ell^* = 8$) & Ex. Match + Random    & 83.0 & 85.4 \\
Fixed ($\ell^* = 8$) & Ex. Match + Discard   & 82.5 & 81.3 \\
Flexible             & Hung. only          &   83.0   &   87.7   \\
Verif. $\to$ Summ. & Ex. Match + Hung. & 69.5	& 83.6 \\
\bottomrule
\end{tabular}
\label{tab:matching_ablation}
\end{table}
\begin{table*}[t]
\centering

\begin{minipage}[t]{0.32\textwidth}
\centering
\caption{Ablation over number of heads selected.}.
\begin{tabular}{c|c}
\toprule
$K$ & \textbf{Exact} \\
\midrule
5  & 87.5 \\
10 & 84.5 \\
15 & 80.0 \\
25 & 81.5 \\
32 & 63.5 \\
\bottomrule
\end{tabular}
\label{tab:k_ablation}
\end{minipage}
\hfill
\begin{minipage}[t]{0.62\textwidth}
\centering
\caption{Ablation over loss components.}
\begin{tabular}{ccc|ccc}
\toprule
$L_{ce}$ & $L_{entropy}$ & $L_{consistency}$ & \textbf{Exact} & \textbf{FitnessQA} & \textbf{ExpertAF} \\
\midrule
$\checkmark$ &  &  & 84.5 & 74.1 & 37.5 \\
$\checkmark$ & $\checkmark$ &  & 77.5 & 77.7 & 37.5 \\
$\checkmark$ &  & $\checkmark$ & 85.0 & 86.8 & 36.8 \\
$\checkmark$ & $\checkmark$ & $\checkmark$ & \textbf{87.5} & \textbf{88.2} & \textbf{38.3} \\
\bottomrule
\end{tabular}
\label{tab:loss_ablation}
\end{minipage}
\end{table*}



We conduct ablation studies to investigate the effectiveness of each component of our method. Table~\ref{tab:matching_ablation} ablates the verification layer, the strategy used to match unmatched heads in $\mathcal{L}_{\text{consistency}}$, and choice of related task. We consider three alternatives to our proposed Exact Match + Hungarian strategy (first row): (1) \textbf{randomly} matching the unmatched head sets from each pathway, (2) \textbf{discarding} unmatched heads entirely, and (3) allowing the verification pathway to \textbf{flexibly} select its visual layer, using the top-$K$ heads from the generation pathway as anchors for matching. Both random assignment (83.0/85.4 accuracy on Exact/FitnessQA) and discarding (82.5/81.3) underperform our method (87.5/88.2), confirming that the quality of head correspondence directly determines the quality of the consistency signal. Discarding might be particularly harmful at $K=5$, where losing even 1-2 unmatched heads leaves too few heads contributing to the attention-based losses, whereas random assignment at least ensures all K heads participate despite the noisy correspondence. The flexible layer strategy also underperforms (83.0/87.7) compared to our method (87.5/88.2), indicating that sharing the same layer $\ell^*$ across pathways is important for producing attention maps that are comparable.

We additionally ablate the choice of the related task. We replace the verification task which is tightly related to the generation task, with a general summarization task (``Summarize this video.''). 
This leads to a significant drop in performance (69.5/83.6) compared to our method, confirming that the complementary task structure is essential. Summarization distributes attention broadly across all frames, and so the consistency signal would penalize the generation pathway for attending selectively to the relevant frames. The comparably smaller drop on FitnessQA may be partly due to the repetitive nature of fitness exercises, where relevant frames could be more broadly distributed across the video, partially overlapping with the diffuse attention of the summarization task.

\textbf{Number of selected heads.} Table~\ref{tab:k_ablation} shows that $K=5$ achieves the best performance (87.5), with accuracy sharply degrading as $K$ increases. We attribute this to the inclusion of non-vision centric heads at larger $K$ values, which introduce noisy attention maps that dilute the consistency signal. 

\textbf{Loss components.} Table~\ref{tab:loss_ablation} ablates the contribution of each loss term. Removing $\mathcal{L}_{consistency}$, while keeping only the regularization losses leads to a significant drop (77.5 on Exact / 77.7 on FitnessQA), confirming that $\mathcal{L}_{consistency}$ is essential to improving understanding capability rather than only enforcing peakiness through the $\mathcal{L}_{entropy}$. Using $\mathcal{L}_{ce}$ with $\mathcal{L}_{consistency}$ alone (row 3, 85.0 / 86.8) already surpasses the finetuned baseline, demonstrating that the consistency signal is the primary driver of improvement. Adding $\mathcal{L}_{entropy}$ (full model, row 4) further improves performance (87.5 / 88.2) and improves from 36.8 to 38.3 on ExpertAF, validating its role in encouraging focused attention distributions.

\section{Conclusion}

We presented a dual-pathway self-consistency framework for video-language model training that improves internal frame localization without requiring any frame-level annotations. By enforcing attentional agreement between a generation pathway and an easier verification pathway, we encourage the model to ground its responses in the same relevant video frames regardless of task, using the consistency between these complementary task-views as a self-supervision signal. We show that Vid-LLMs tend to misallocate attention toward irrelevant frames by observing improvements through our self-consistency training across diverse sports understanding benchmarks. We hope this work motivates further exploration of internal attention structure and complementary task-views as a source of self-supervision in video-language models.
\bibliographystyle{splncs04}
\bibliography{main}

\title{Supplemental Section for Learning Consistent Temporal Grounding between Related Tasks in Sports Coaching} 

\titlerunning{Supplemental Section}
\author{Arushi Rai \and
Adriana Kovashka
}
\authorrunning{A.~Rai and A.~Kovashka}

\institute{University of Pittsburgh, Pittsburgh PA 15260, USA \\
\email{arr159@pitt.edu, kovashka@cs.pitt.edu}}



\maketitle


This supplementary material provides additional details and experiments to
complement the main paper.
Sec.~\ref{supp:vnsr} gives the full definition of the Visual Non-Sink Ratio \cite{kang2025see} used for adaptive head selection (referenced in Sec.~3.3 of the main paper). Sec.~\ref{supp:attn_redist} describes and illustrates the proportional attention
redistribution strategy used in our proof-of-concept analysis
(referenced in Sec.~4.2). Sec.~\ref{supp:fitnessqa_mc} provides the prompt and conversion examples for transforming FitnessQA open-ended annotations into multiple-choice questions (referenced in Sec.~4). 
Sec.~\ref{supp:stopgrad} presents an ablation on the direction of the stop-gradient operator in $\mathcal{L}_{\text{consistency}}$.
Finally, Sec.~\ref{supp:qualitative} provides qualitative examples of 
generated coaching feedback.

\section{Visual Non-Sink Ratio}
\label{supp:vnsr}
We provide the definition of Visual Non-Sink Ratio\cite{kang2025see} in this section.
Given a hidden state $\mathbf{x}^{\ell-1}_j \in \mathbb{R}^D$ of the $j$-th token at layer $\ell$, the sink dimension value is:
    \begin{equation}
        \phi(\mathbf{x}^{\ell-1}_j) = \max_{\check{d} \in \mathcal{D}_{\text{sink}}} 
        \left| \frac{\mathbf{x}^{\ell-1}_j[\check{d}]}{\sqrt{\frac{1}{D}
        \sum_{d=1}^{D} \mathbf{x}^{\ell-1}_j[d]^2}} \right|,
    \end{equation}
    where $\mathcal{D}_{\text{sink}}$ is the set of sink dimensions of the base language model. The set of visual sink token indices at layer $\ell$ is then $\check{\mathcal{I}}^{\ell}_{\text{vis}} = \check{\mathcal{I}}^{\ell} \cap \mathcal{I}_{\text{vis}}$, where $\check{\mathcal{I}}^{\ell} = \{ j \in 
    \mathcal{I} \mid \phi(\mathbf{x}^{\ell-1}_j) \geq \tau \}$ represents the set of sink token indices, $\mathcal{I}_{\text{vis}}$ the set of visual token indices, and visual non-sink tokens form the remainder $\mathcal{I}_{\text{vis}} \setminus \check{\mathcal{I}}^{\ell}_{\text{vis}}$.
    The Visual Non-Sink Ratio is defined for each head $i$ as:
    \begin{equation}
        \mathrm{VNSR}_i =
    \frac{
    \sum\limits_{j \in \mathcal{I}_{\text{vis}} \setminus \check{\mathcal{I}}_{\text{vis}}}
    A_{i,j}
    }{
    \sum\limits_{j \in \mathcal{I}_{\text{vis}}}
    A_{i,j}
    },
    \end{equation}
    where $A\in \mathbb{R}^{H \times N}$ is the attention map from the last prompt token and $H$ is the number of heads and $N$ is the number of tokens.
\section{Proportional Attention Redistribution}
\label{supp:attn_redist}

We conducted an analysis in Sec.~4.1 of the main paper where we identified a temporal-grounding specific layer and head to which we apply our method. But first, we validated this by redistributing visual attention based on the ground-truth keyframe annotations. Since modern video-language models~\cite{cho2025PerceptionLM, wang2025internvideo, qwen25} represent each frame with multiple visual tokens, we choose to proportionally redistribute attention based on the original vision proportions.

Formally, let $\mathcal{F}$ denote the set of all sampled frames and
$\mathcal{K} \subseteq \mathcal{F}$ the subset of ground-truth keyframes.
Each frame $f$ is associated with a contiguous block of visual tokens
$\mathcal{I}_{f} \subset \mathcal{I}_{\text{vis}}$.
Given the original attention distribution
$\mathbf{A}_{i} \in \mathbb{R}^{N_{\text{vis}}}$ for head $i$, the sum of the attention over all the visual tokens, $S_i=\sum_{j \in \mathcal{I}_{\text{vis}}} A_{i,j}$, and the sum of the attention over keyframes, $S_i^{\mathcal{K}}=\sum_{j \in \bigcup_{f \in \mathcal{K}} \mathcal{I}_{f}} A_{i,j}$, the
proportionally redistributed map $\tilde{\mathbf{A}}_{i}$ is:

\begin{equation}
    \tilde{A}_{i,j}
    =
    \begin{cases}
        A_{i,j} \cdot \dfrac{S_i}{S_i^{\mathcal{K}}}
          & j \in \bigcup_{f \in \mathcal{K}} \mathcal{I}_{f}, \\[8pt]
        0 & \text{otherwise}.
    \end{cases}
\end{equation}
The scaling factor $S_i / S_i^{\mathcal{K}}$ rescales keyframe token
attention so that the total visual attention mass is preserved
(i.e.\ $\sum_{j \in \mathcal{I}_{\text{vis}}} \tilde{A}_{i,j}=S_i$),
while non-keyframe visual tokens are zeroed out.
The relative attention weights \emph{within} a keyframe are unchanged. Attention to text and system tokens is left completely untouched. 

\begin{figure}
    \centering
    \includegraphics[width=0.5\linewidth]{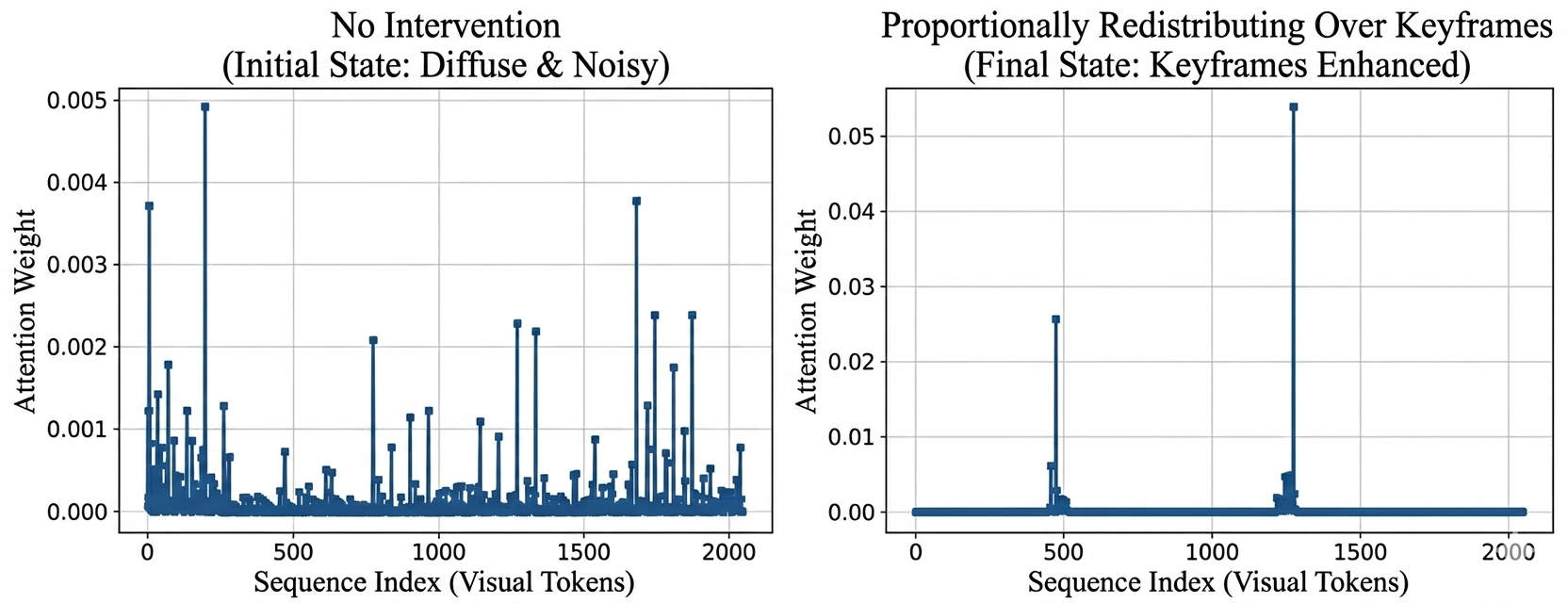}
    \caption{\textbf{Attention redistribution strategies.} Left: The initial attention distribution is \textit{diffuse} and noisy across the entire visual sequence. Right: \textit{Proportional redistribution} concentrates the attention weights onto specific keyframe tokens while preserving their original relative importance.}    
    \label{fig:redistribution}
\end{figure}

Figure~\ref{fig:redistribution} illustrates this method.

\section{FitnessQA Multiple-Choice Conversion}
\label{supp:fitnessqa_mc}

FitnessQA~\cite{panchal2024say} was released with open-ended reference answers, which require either an LLM-as-a-judge or imprecise semantic similarity metrics to evaluate automatically. Converting to MCQ not only enables straightforward accuracy-based evaluation consistent with Exact~\cite{yi2025exact}, but also provides a more natural and discriminative format for questions concerning degree or magnitude (e.g. stance width, squat depth), where graded options better capture the range of possible movement quality than a single open-ended response. We convert each set of fine-grained questions, corresponding answers, and exercise information into a four-option multiple-choice question using an LLM. The conversion prompt is given below, followed by examples in Table~\ref{tab:supp_fitnessqa_mc_examples}.

\begin{tcolorbox}[promptbox, title=FitnessQA Multiple-Choice Conversion Prompt]

\textbf{You are given fine-grained Yes/No coaching QA pairs about a user's movement.}

\medskip
\textbf{TASK} From the input, derive EXACTLY 3 multiple-choice questions (MCQs) in JSON format. Each MCQ must include:
\begin{itemize}[leftmargin=1.5em, itemsep=2pt]
  \item \texttt{"question"}: a clear, standalone question derived from ONE of the input queries
  \item \texttt{"correct\_answer"}: the correct answer option text
  \item \texttt{"distractor\_answers"}: an array of 3 plausible but incorrect answer option texts
\end{itemize}

\medskip
\textbf{OUTPUT FORMAT} Return ONLY valid JSON (no markdown, no commentary), shaped exactly like:
\begin{verbatim}
{
  "mcq": [
    {
      "question": "...",
      "correct_answer": "...",
      "distractor_answers": ["...", "...", "..."]
    },
    ...
  ]
}
\end{verbatim}

\medskip
\textbf{RULES}
\begin{enumerate}[leftmargin=1.5em, itemsep=2pt]
  \item Choose 4 input items that are diverse (i.e.\ their correct answers are a mix of `yes'/`no' and positive/negative feedback).
  \item Convert the original Yes/No query into a multiple-choice form by switching to a categorical question where possible.
  \begin{itemize}
    \item If query is about depth/angle: use options like \texttt{["Above 90°", "About 90°", "Below 90°", "Cannot determine"]}.
    \item If query is about width: use options like \texttt{["Too narrow", "Appropriate", "Too wide", "Asymmetrical/Uneven"]}.
  \end{itemize}
  \item The \texttt{"correct\_answer"} MUST match the exact wording of the provided response without yes/no text.
  \item Distractors MUST be plausible, mutually distinct, and not synonyms of the correct answer.
  \item Keep wording consistent and concise; do not introduce new body parts or claims not supported by the input.
  \item If the input response is grammatically incorrect, preserve meaning but you may fix grammar in the MCQ options.
  \item Ensure each MCQ can be answered without needing the other questions.
  \item Include the exercise type in the question.
\end{enumerate}

\medskip
\textbf{INPUT}

\end{tcolorbox}

\begin{table}[h]
\centering
\small
\caption{%
  \textbf{FitnessQA multiple-choice conversion examples.}
  The correct option (marked \checkmark) is a paraphrase of the original
  open-ended reference answer; distractors are plausible but incorrect
  alternatives for the same exercise.
}
\label{tab:supp_fitnessqa_mc_examples}
\setlength{\tabcolsep}{4pt}
\begin{tabular}{lp{5.0cm}p{5.0cm}}
\toprule
 & \textbf{Example 1} & \textbf{Example 2} \\

\textbf{Question}
  & Is the user's movement sufficient during shoulder gators?
  & While performing plank taps on their knees, are the user's hips elevated? \\[4pt]
\textbf{Reference}
  & Yes, the movement is insufficient.
  & No, the user's hips are not elevated. \\[4pt]
\textbf{A}
  & The movement is excessive.
  & The user's hips are significantly elevated. \\
\textbf{B}
  & \textbf{The movement is insufficient.}~\checkmark
  & The user's hips are slightly elevated. \\
\textbf{C}
  & The movement is appropriate.
  & \textbf{The user's hips are not elevated.}~\checkmark \\
\textbf{D}
  & The movement is too slow.
  & The user's hips are at an appropriate height. \\
\bottomrule
\end{tabular}
\end{table}

\section{Stop-Gradient Direction Ablation}
\label{supp:stopgrad}

In our method, the stop-gradient operator in $\mathcal{L}_{\text{consistency}}$
is applied to the \emph{verification} pathway, treating it as a fixed target
toward which the generation pathway is trained (Eq.~4 of the main paper).
This design choice is motivated by two factors:
(i)~verification is a simpler task, so its attention maps provide a more
reliable target signal; and
(ii)~training collapse is avoided by preventing gradients from flowing
through both pathways simultaneously, similar to many self-supervised work with dual-pathways\cite{grill2020bootstrap, he2020momentum, caron2021emerging}.

To validate this choice, we compare against the \emph{reversed} direction,
in which the stop-gradient is applied to the \emph{generation} pathway instead:
\begin{equation}
    \mathcal{L}_{\text{consistency}}^{\text{rev}}
    = \frac{1}{K} \sum_{i=1}^{K}
      D_{\mathrm{KL}}\!\Bigl(
        \mathrm{sg}\!\left(\mathbf{A}^{\text{gen}}_i\right)
        \;\Big\|\;
        \mathbf{A}^{\text{ver}}_i
      \Bigr).
\end{equation}
and additionally, $L_{entropy}$ will also be applied to attention maps from the verification pathway only. We found training with $L_{ce}$ on only the verification pathway caused the model to output only verification-style output (i.e. ``yes'', ``no'') on question-answering (not feedback generation), so we also have a setting where $\mathcal{L}_{ce}$ is omitted from the verification pathway.

Results are shown in Table~\ref{tab:supp_stopgrad}. We observe severely degraded performance across all benchmarks for both reversed-direction settings. In the setting where $\mathcal{L}_{ce}$ is omitted from the verification pathway, either none or very few of the generated answers matched the required format on Exact and FitnessQA, yielding near-zero accuracy. On ExpertAF, the rescaled BERTScore is $-4.0$, indicating that the generated feedback is worse than a random baseline, consistent with the model producing degenerate outputs in the absence of $\mathcal{L}_{ce}$ enforcement. When $\mathcal{L}_{ce}$ is additionally applied to the verification pathway, performance partially recovers on ExpertAF ($16.1$), as the cross-entropy loss provides some generative supervision; however, accuracy on Exact and FitnessQA remains at or near zero, as the model collapses to verification-style outputs (\ie ``yes'', ``no'') rather than producing answers in the required format. 

These failures modes to a fundamental problem: when the evaluation task is generative, the model must be trained on generation via $\mathcal{L}_{ce}$. Without it on the generation pathway, LoRA fine-tuning causes the model to forget how to produce well-formed generative outputs, collapsing either to verification-style responses or to incoherent outputs. Our original design avoids this problem entirely, as $\mathcal{L}_{ce}$ is always applied to the generation pathway, preventing such forgetting. We note that an ablation applying $\mathcal{L}_{ce}$ exclusively to the generation pathway and $L_{\text{entropy}}$/$L_{\text{consistency}}$ to the verification pathway could solve this problem, but was not feasible under our memory constraints.

\begin{table}[t]
\centering
\small
\caption{%
  \textbf{Stop-gradient direction ablation.}
  ``Ours (ver.\ fixed)'' applies \texttt{sg} to the verification pathway
  (Eq.~4 of the main paper).
  ``Gen.\ fixed'' reverses the direction, fixing the generation pathway
  and training the verification pathway to match it. On Exact, none of the generated answers matched the format. 
}
\label{tab:supp_stopgrad}
\begin{tabular}{lccc}
\toprule
\textbf{SG Direction} & \textbf{Exact} & \textbf{FitnessQA} & \textbf{ExpertAF} \\
\midrule
Ours (verification fixed) & \textbf{87.5} & \textbf{88.2} & \textbf{38.4} \\
Generation fixed (no $L_{ce}$ on verif)          & 0.0          & 2.7          & -4.0           \\
Generation fixed (w/ $L_{ce}$ on verif)          & 1.0          & 0.0          & 16.1          \\
\bottomrule
\end{tabular}
\end{table}

\section{Qualitative Examples}
\label{supp:qualitative}





\begin{figure}[]
    \centering
    \begin{tabular}{cp{0.85\linewidth}}
        \rotatebox{90}{\small Soccer} &
        \includegraphics[width=\linewidth]{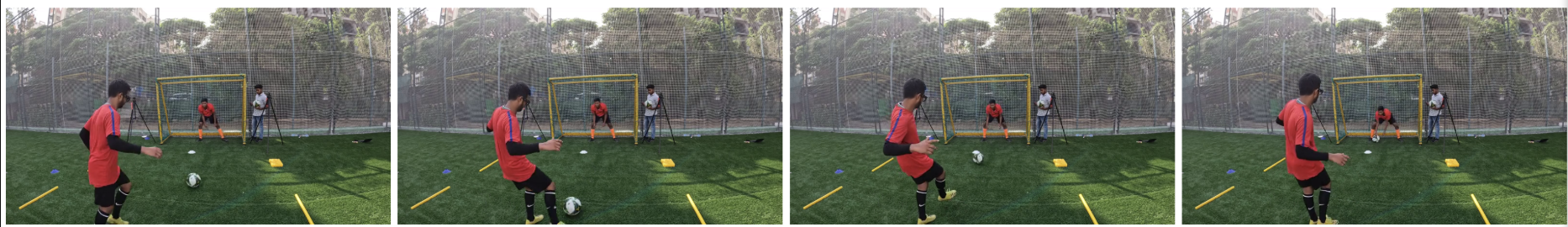} \\[4pt]
        \rotatebox{90}{\small Climbing} &
        \includegraphics[width=\linewidth]{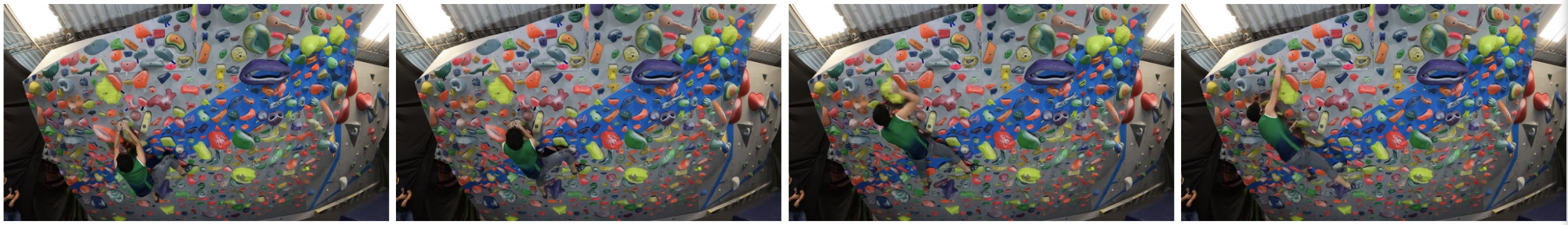} \\
    \end{tabular}
    \caption{Selected frames from the ExpertAF clips in 
    Table~\ref{tab:supp_feedback_qual}.}
    \label{fig:supp_feedback_frames}
\end{figure}
\begin{table}[]
\centering
\small
\caption{%
  Qualitative feedback examples on ExpertAF.
}
\label{tab:supp_feedback_qual}
\setlength{\tabcolsep}{4pt}
\begin{tabular}{p{1.6cm}p{3.0cm}p{3.0cm}p{3.0cm}}
\toprule
 Sport& \textbf{Reference} & \textbf{Finetuned baseline} & \textbf{Ours} \\
\midrule
Soccer
  & The player's kicking technique needs improvement, especially the follow-through and body alignment to maximize power, control, and accuracy.
  & The player's footwork and body position are affecting the quality of the pass.
  &  The player shows good approach technique to the ball, but needs to improve follow-through and body positioning for maximum power and accuracy. \\[6pt]
Climbing
  &  The climber is attempting a dynamic move that involves cutting his feet loose and supporting his weight with his arms, which is a challenging and uncertain approach.
  & The climber's footwork is not ideal; he should use his left foot to drive upward and bring his right foot back onto the hold.
  & The climber successfully executed a dynamic move, starting from a good position with effective foot placement to reach the next hold.  \\
\bottomrule
\end{tabular}
\end{table}
Table~\ref{tab:supp_feedback_qual} shows coaching feedback generated by the finetuned baseline and our method on select ExpertAF clips. Qualitative evaluation of open-ended feedback is inherently noisy, and these examples should be interpreted cautiously. That said, our method more frequently captures the key action  present in the reference (e.g.\ ``follow-through'' in the soccer clip and identifying the dynamic move in the climbing clip), whereas the baseline identified less specific actions (``footwork'', instead of the specific component of footwork like ``follow-through'') or problems that were not mentioned in the reference (``footwork'', row 2). We attribute this to better temporal grounding: attending to the specific frames where the key action occurs makes it more likely that the generated feedback references that action. However, the model still struggles to reason about the \emph{impact} of the observed action or if it was the optimal move. This is shown by the climbing clip in row 2, where the model states that the move is effective whereas the reference describes it as challenging and risky. This limitation suggests that improvements to temporal grounding are insufficient to address higher-order reasoning about action consequences. We posit that such capabilities may require subsequent post-training on fine-grained, domain-specific knowledge related to action impact and optimality. 


\end{document}